\documentclass[11pt]{article}

\usepackage[final]{acl}

\usepackage{times}
\usepackage{latexsym}

\usepackage[T1]{fontenc}

\usepackage[utf8]{inputenc}

\usepackage{microtype}

\usepackage{inconsolata}

\usepackage{graphicx}

\usepackage{amsmath}
\usepackage{amssymb}
\usepackage{booktabs}
\usepackage{multirow}
\usepackage{stfloats}
\usepackage{cuted}
\usepackage{caption}

\newcommand{\method}{PASM} 

\title{PASM: Population Adaptive Symbolic Mixture-of-Experts Model for Cross-location Hurricane Evacuation Decision Prediction}

\author{
  Xiao Qian \\
  Dept. of Civil, Environmental,\\
  and Construction Engineering \\
  University of Delaware, \\
  Newark, DE, USA \\
  \texttt{qxiao@udel.edu}
  \And
  Shangjia Dong \\
  Dept. of Civil, Environmental,\\
  and Construction Engineering \\
  University of Delaware, \\
  Newark, DE, USA \\
  \texttt{sjdong@udel.edu}
}

\begin{document}
\maketitle


\begin{abstract}

Accurate prediction of evacuation behavior is critical for disaster preparedness, yet models trained in one region often fail elsewhere. Using a multi-state hurricane evacuation survey, we show this failure goes beyond feature distribution shift: households with similar characteristics follow systematically different decision patterns across states. As a result, single global models overfit dominant responses, misrepresent vulnerable subpopulations, and generalize poorly across locations. We propose Population-Adaptive Symbolic Mixture-of-Experts (PASM), which pairs large language model guided symbolic regression with a mixture-of-experts architecture. PASM discovers human-readable closed-form decision rules, specializes them to data-driven subpopulations, and routes each input to the appropriate expert at inference time. On Hurricanes Harvey and Irma data, transferring from Florida and Texas to Georgia with 100 calibration samples, PASM achieves a Matthews correlation coefficient of 0.607, compared to XGBoost (0.404), TabPFN (0.333), GPT-5-mini (0.434), and meta-learning baselines MAML and Prototypical Networks (MCC $\leq$ 0.346). The routing mechanism assigns distinct formula archetypes to subpopulations, so the resulting behavioral profiles are directly interpretable. A fairness audit across four demographic axes finds no statistically significant disparities after Bonferroni correction. PASM closes more than half the cross-location generalization gap while keeping decision rules transparent enough for real-world emergency planning.

\end{abstract}

\section{Introduction}

Disasters force communities to make evacuation decisions under extreme time pressure and with life-or-death consequences. Predicting who will evacuate is essential for emergency managers to allocate resources, coordinate evacuations, and prioritize rescue efforts. While frameworks such as the Protective Action Decision Model (PADM) provide a conceptual foundation \cite{lindell2012protective}, real households are highly heterogeneous, varying in mobility, income, language access, caregiving responsibilities, and medical dependence \cite{perry2007disaster, cutter2003social}. One-size-fits-all models risk overlooking these vulnerable populations, who are often treated as statistical outliers yet face the highest risk \cite{fothergill2004poverty}. The consequences are well documented, from language barriers during Hurricane Katrina \cite{elder2007minorities} to evacuation plans that implicitly assumed universal car ownership \cite{litman2006lessons}.

Accurately predicting evacuation behavior is difficult, particularly when generalizing these models across regions. Social, economic, and cultural differences lead to substantial variation in decision-making, and models calibrated in one context often fail elsewhere. For example, wildfire studies identify both “evacuation-keen” and “evacuation-reluctant” subpopulations, implying that any single global model will misrepresent at least one group \cite{Wong2023}. Ignoring this heterogeneity results in mistargeted warnings, inefficient resource allocation, and inequitable outcomes.

\begin{figure*}[t]
\centering
\includegraphics[width=0.9\linewidth]{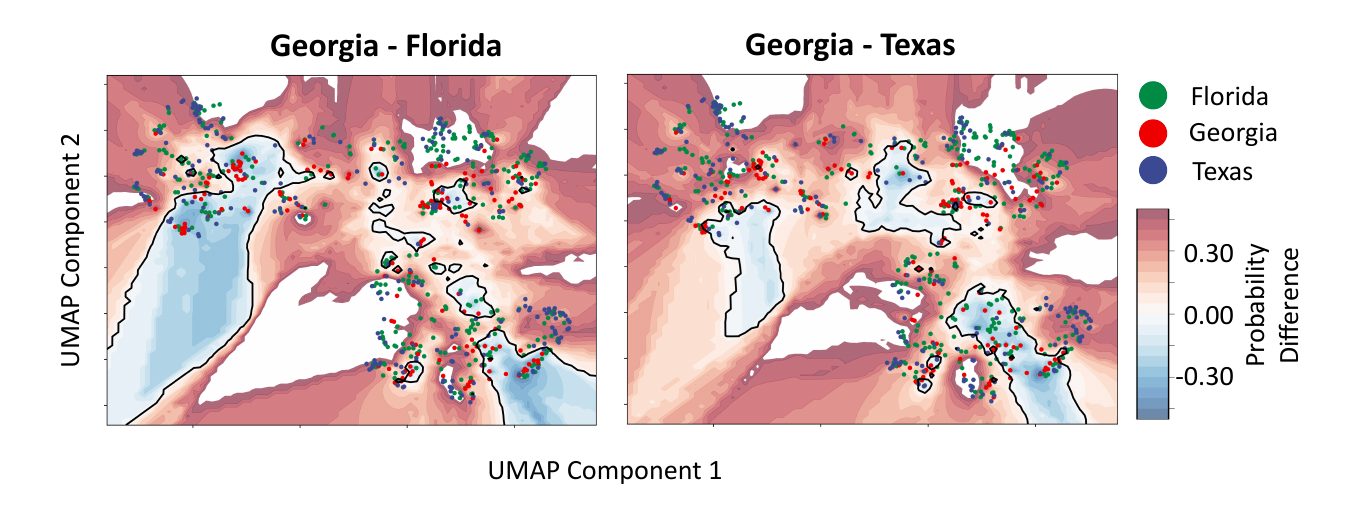}
\caption{Evacuation probability difference in UMAP (Uniform Manifold Approximation and Projection) space: $P(\text{Evac}|\text{Georgia}) - P(\text{Evac}|\text{Florida})$ (left) and $P(\text{Evac}|\text{Georgia}) - P(\text{Evac}|\text{Texas})$ (right).}
\label{fig:state_effect_heatmap}
\end{figure*}

\paragraph{Challenges} We examine the model transferability on several state-of-the-art models regarding how well they work when trained in one state and applied to others. We use a household survey dataset collected after Hurricanes Harvey and Irma \cite{goodie2019experience, goodie2019data}, containing 822 anonymized respondents from Texas (Harvey) and Florida and Georgia (Irma) in the United States, containing household evacuation decisions and associated socioeconomic, cognitive, and experiential factors. We train all models using data from Florida and then test them on Florida, Texas, and Georgia. The results in Table~\ref{tab:cross_state_mcc} show a clear pattern: all models perform much worse when applied to Georgia. However, the performance remains largely unchanged when transferred between Florida and Texas. This asymmetric pattern of Florida-Texas compatibility, but a pronounced gap with Georgia, suggests that Georgia residents behave differently in evacuation decisions.

To better understand why this happens, we visualize how evacuation decisions differ across states for people with the same characteristics in Figure~\ref{fig:state_effect_heatmap}. Each point in the visualization represents individuals with identical profiles (same age, income, risk perception, etc.). If people across states made decisions in the same way, the differences would be close to zero everywhere. Instead, we see strong heterogeneity: large red regions (differences up to 0.30-0.45). In many cases (i.e., dominance of red regions), people in Georgia are much more likely to evacuate than similar individuals in Florida or Texas; blue regions indicate the opposite is true. Thus, identical factors can exert opposite effects across states. As a result, a single global model cannot reconcile conflicting behavior across states. 

There is also substantial intra-state behavioral heterogeneity, limiting the effectiveness of a single global model. To examine this, we compare three settings: (i) training and testing on Florida (FL $\rightarrow$ FL), (ii) training on Florida and testing on Georgia (FL$\rightarrow$ GA), and (iii) training and testing on Georgia (GA$ \rightarrow$ GA). If Georgia were behaviorally homogeneous, GA$\rightarrow$ GA would be expected to outperform FL$\rightarrow$ GA. Table~\ref{tab:intra_state_heterogeneity} reveals the opposite. Models trained on Georgia perform worse on Georgia test data than models trained on Florida. This counterintuitive result reflects intra-state heterogeneity under limited data. With only 100 training samples, Georgia-trained models overfit to the specific subgroups observed, failing to generalize across the state’s diverse behavioral patterns. In contrast, Florida’s broader training distribution yields decision rules that transfer more robustly, despite being out-of-state.

In all, we show that behavioral heterogeneity operates at multiple levels, both across and within states. Any single global model will favor dominant patterns and systematically underperform for minority subpopulations. This motivates our Mixture-of-Experts framework, which explicitly learns and combines multiple behavioral regimes rather than forcing a one-size-fits-all predictor.

\paragraph{Contribution} We address the challenges by integrating symbolic regression and Mixture-of-Experts (MoE) modeling. Symbolic regression yields transparent decision rules with strong extrapolation power, and recent LLM-guided methods such as LaSR and DrSR outperform black-box models under distribution shift \cite{Cranmer2020, grayeli2024lasr, wang2025drsr}. However, a single symbolic equation is insufficient: rules learned in one state generalize poorly to others. MoE architectures provide a remedy by combining specialized models with learned routing, enabling shared structure while isolating conflicting patterns \cite{tian2023decompose, zhao2025learning}. We thus propose Population-Adaptive Symbolic MoE (PASM), which couples LLM-guided symbolic experts with a learned controller to achieve an interpretable and robust evacuation model. PASM is distinct from the PADM framework mentioned earlier: whereas PADM provides a theory-driven model designed by domain experts, decomposing protective action decisions into sequential cognitive stages defined a priori. PASM, by contrast, is entirely data-driven: it uses symbolic regression to discover decision rules directly from observations, bypassing the need for expert-specified cognitive assumptions, and routes inputs through multiple symbolic experts to capture heterogeneous behavioral patterns.

\section{Related Work}

\subsection{Evacuation Decision Prediction for Diverse Population}

Evacuation research is grounded in the Protective Action Decision Model (PADM), which frames evacuation as a sequence of cognitive stages influenced by environmental cues, social signals, and official warnings \cite{lindell2012protective, huang2016who}. Empirical studies have largely operationalized PADM using logistic regression and discrete choice models to estimate evacuation likelihoods based on factors such as housing type, pet ownership, and prior experience \cite{hasan2011behavioral, dash2007evacuation, goodie2019experience}. While interpretable, these models assume a homogeneous decision process and perform poorly when transferred across regions or events.

Machine learning approaches have improved predictive accuracy by capturing nonlinear interactions \cite{sun2024predicting}, yet they continue to suffer from a persistent transfer gap: models trained on one disaster or region often generalize poorly to others due to latent spatial and temporal heterogeneity \cite{morss2016effects}. This limitation is also closely tied to concerns of algorithmic fairness, as global models tend to favor majority behaviors and systematically underperform for vulnerable subpopulations, such as low-income households without private vehicles \cite{gevaert2021fairnessdrm}.

Recent work has explored personalized and modular approaches. For example, ATHENA \cite{zhao2025personalized} uses LLMs to infer individualized utility functions for decision-making, but relies on rule-based subgrouping and per-instance optimization, limiting scalability and control over bias. Our work differs by combining data-driven subpopulation discovery, symbolic regression for interpretable decision rules, and a learned Mixture-of-Experts architecture to enable transferable and population-adaptive evacuation modeling.

\subsection{LLM-Augmented Symbolic Regression}

Symbolic Regression (SR) aims to discover explicit mathematical expressions that explain observed data, jointly searching over equation structure and parameters rather than assuming a fixed functional form. Its inherent interpretability makes SR suitable for high-stakes decision modeling, where black-box neural networks are often viewed with skepticism by policymakers \citep{schmidt2009distilling}. Traditional SR methods have largely relied on genetic programming (GP), as implemented in tools such as \textit{Eureqa} and \textit{gplearn}. While effective in low-dimensional settings, GP-based approaches scale poorly and yield overly complex or physically implausible expressions \citep{petersen2019deep}.

Recent advances in large language models (LLMs) have revitalized SR by introducing strong priors over plausible functional forms. LLM-SR \citep{shojaee2025llmsr} leverages pretrained language models to propose equation skeletons, substantially reducing search complexity and sample requirements. LaSR \citep{grayeli2024lasr} further improves efficiency by using LLMs to learn and reuse symbolic concepts, while DrSR \citep{wang2025drsr} incorporates dual reasoning to iteratively refine symbolic hypotheses based on data feedback. They all demonstrated strong performance in recovering governing equations in physics and biology.

However, most LLM-augmented SR methods are designed to identify a single global equation, an assumption that does not hold in social systems where behavior varies across subpopulations. In evacuation modeling, no universal decision law exists. Our proposed PASM framework adapts LLM-guided symbolic regression to this setting by generating a diverse set of interpretable behavioral heuristics and embedding them within a population-adaptive Mixture-of-Experts framework.

\begin{figure*}[t]
\centering
\includegraphics[width=\textwidth]{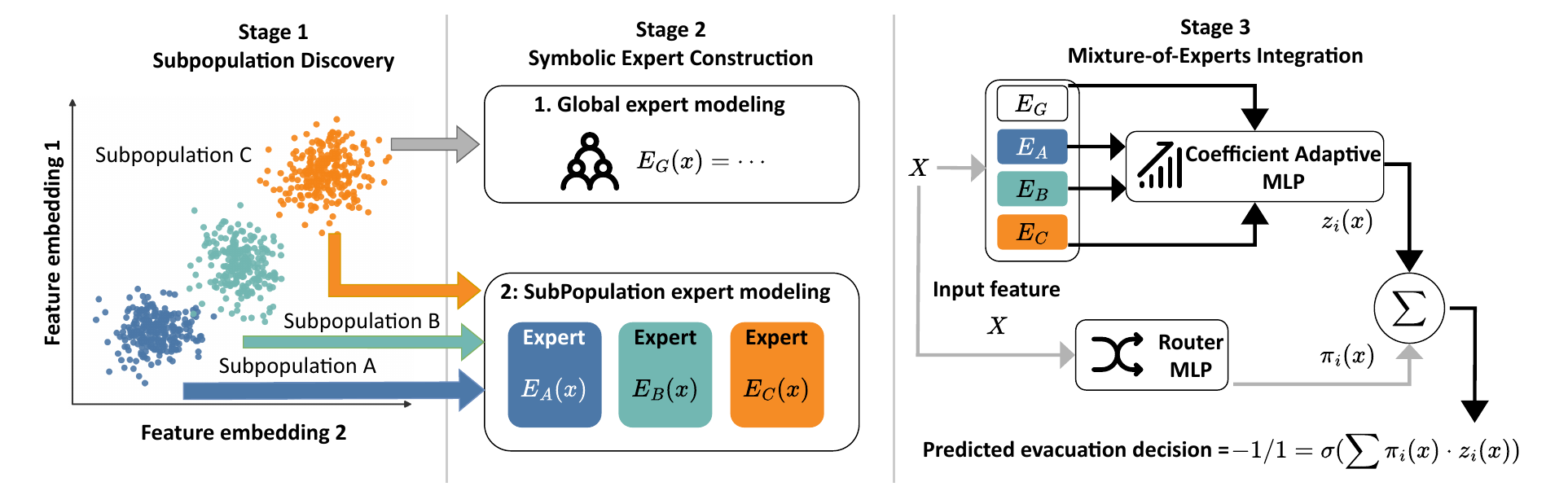}
\caption{Overview of the PASM framework.}
\label{fig:framework}
\end{figure*}

\subsection{Mixture-of-Experts and Multi-Task Learning}

While SR provides interpretable equations for modeling individual decision rules, a single global formula is often insufficient to capture heterogeneous behavior across subpopulations. Mixture-of-Experts (MoE) architectures offer a natural solution to this challenge. Originally proposed by \citet{jacobs1991adaptive}, MoE trains multiple specialized sub-models (``experts") along with a gating network (``router") that assigns inputs to the most appropriate expert. This modularity allows different experts to capture distinct patterns within heterogeneous populations while sharing common information where appropriate.

MoE has gained a lot of interest in deep learning, especially with sparse transformer variants such as Mixtral \citep{jiang2024mixtral} and DeepSeekMoE \citep{dai2024deepseekmoe}, which scale efficiently to large models. A key challenge in MoE, particularly in transfer and multi-task settings, is managing interactions between shared and specialized parameters and avoiding ``negative transfer," where optimizing one expert conflicts with others \citep{standley2020tasks, yu2020gradient}. Approaches like MoDULA \citep{ma2024modula} address this by separating domain-specific experts from a universal expert and employing staged training to stabilize learning. Similar principles have been applied in robotics and reinforcement learning: DT2GS \citep{tian2023decompose} decomposes multi-agent tasks into sub-tasks, and M3W \citep{zhao2025learning} routes diverse dynamics to specialized experts, improving generalization across contexts.

Adapting MoE to SR introduces additional advantages. Recent work, such as Symbolic-MoE \citep{chen2025symbolic}, routes queries to different LLMs, forming an ensemble of black-box models. In contrast, our PASM framework integrates SR with a learned MoE controller, enabling multiple interpretable symbolic experts to model distinct behavioral regimes within the population. The gating network adaptively selects or combines experts for each individual based on their features, allowing the model to capture heterogeneous evacuation behaviors while maintaining interpretability. This design effectively extends the benefits of MoE to population-adaptive symbolic modeling, bridging the gap between interpretable rule discovery and scalable, heterogeneous behavioral prediction.

\section{Method} \label{sec:method}

The PASM framework models a function $F(x)$ that maps household and situational features $x \in \mathbb{R}^d$ to a binary evacuation decision $y \in \{0, 1\}$. To capture population heterogeneity and account for cross-location distribution shifts, $F(x)$ is structured as a Mixture-of-Experts, where each expert $E_k(x)$ is an interpretable symbolic expression tailored to a subset of the population.

In Stage 1 (Subpopulation Discovery), we uncover latent subpopulations by embedding household features into a low-dimensional space and applying unsupervised clustering. Using UMAP \cite{mcinnes2018umap} for dimensionality reduction and HDBSCAN \cite{mcinnes2017hdbscan} for density-based clustering, we partition the training data into coherent subgroups with distinct evacuation behaviors without pre-specifying the number of clusters.

In Stage 2 (Symbolic Expert Construction), we build a library of interpretable symbolic experts by first fitting a global model $E_G(x)$ to capture population-level patterns, then training cluster-specific experts to model subpopulation-specific decision logic. All experts are learned using LaSR \cite{grayeli2024lasr}, an LLM-guided symbolic regression framework that discovers compact and interpretable expressions via concept abstraction.

In Stage 3 (Mixture-of-Experts Integration), the symbolic experts are composed through a learned MoE controller. A Router MLP outputs a distribution
$\pi (x)$ over experts, indicating their relevance for the individual. In parallel, a Coefficient-Adaptive MLP calibrates each expert’s internal coefficients. The final evacuation probability is obtained by applying a sigmoid to the weighted sum of calibrated expert logits. All components are trained jointly end-to-end, improving stability and robustness under distribution shift. Figure~\ref{fig:framework} illustrates the complete pipeline.

\subsection{Symbolic Mixture-of-Experts with Joint Coefficient Adaptation}

Our preliminary experiments show that naively averaging symbolic expert outputs transfers poorly on another state. Expert relevance varies across households: while the global expert suffices for some, others are better explained by subpopulation-specific formulas. Moreover, coefficients learned in one subpopulation do not transfer reliably to another, as identical feature labels (e.g., high income or high risk perception) can have different contextual meanings across regions. For example, a high-income household in rural Georgia faces different constraints than one in Miami. This can lead to errors when coefficients are shared naively.

To address this, PASM employs a learnable routing mechanism that dynamically composes symbolic experts based on input features. Given a feature vector $\mathbf{x} \in \mathbb{R}^d$, the router is a multi-layer perceptron that outputs a probability distribution over the expert library:
\begin{equation}
\boldsymbol{\pi}(\mathbf{x}) = \mathrm{softmax}\left(\mathrm{MLP}_{\text{router}}(\mathbf{x}) / \tau\right) \in \Delta^{M-1},
\end{equation}
where $M$ is the total number of experts (one global expert and $K$ subpopulation-specific experts), and $\tau > 0$ is a temperature parameter. We anneal $\tau$ from a higher initial value $\tau_{\text{init}}$ to a lower final value $\tau_{\text{final}}$ during early training to encourage exploration before converging to sharper routing decisions.

Each symbolic expert $E_m$ produces a scalar logit $z_m(\mathbf{x}; \boldsymbol{\theta}_m)$ reflecting its confidence that household $\mathbf{x}$ will evacuate. To accommodate cross-population differences in scale and interpretation, we apply a learnable affine calibration to each expert:
\begin{equation}
g_m(\mathbf{x}) = \gamma_m \cdot z_m(\mathbf{x}; \boldsymbol{\theta}_m) + \beta_m,
\end{equation}
where $\gamma_m$ and $\beta_m$ are expert-specific scale and bias parameters. The final mixture logit is computed as
\begin{equation}
\hat{z}(\mathbf{x}) = \sum_{m=1}^{M} \pi_m(\mathbf{x}) \cdot g_m(\mathbf{x}).
\end{equation}
and the predicted evacuation probability is $\hat{p}(\mathbf{x}) = \sigma(\hat{z}(\mathbf{x}))$, with $\sigma(\cdot)$ denotes the sigmoid function. Each symbolic expert computes a real-valued output by numerically evaluating its closed-form formula on the input features. A sigmoid function then converts this output to an evacuation probability. The coefficient adaptation network scales expert outputs before the router combines them via learned mixture weights, allowing the same symbolic structure to adapt across heterogeneous subpopulations.

\paragraph{Coefficient Adaptive Network.}
Beyond expert routing, we allow the internal coefficients of symbolic formulas to adapt to the input. For example, in a rule $z(\mathbf{x}) = \theta_1 \cdot x_{\text{wind}} - \theta_2$, the threshold $\theta_2$ reflects risk tolerance and should vary with housing characteristics (e.g., reinforced concrete vs. mobile homes). Rather than defining separate equations, we model coefficients as input-dependent:
\begin{equation}
\boldsymbol{\theta}_m(\mathbf{x}) = \mathrm{MLP}_{\text{coeff},m}(\mathbf{x}).
\end{equation}

The coefficient network shares a common feature backbone across experts while using expert-specific output heads. This enables a single symbolic structure to generalize across heterogeneous subpopulations by modulating its parameters contextually. Jointly adapting both routing weights and coefficients mitigates negative interference across subgroups and allows symbolic rules to flexibly adjust to household-level characteristics.

\paragraph{Joint Optimization and Regularization.}
Unlike staged MoE pipelines that freeze experts during router training \cite{ma2024modula}, PASM is trained end-to-end. We jointly optimize the router parameters $\phi$, symbolic coefficients $\{\boldsymbol{\theta}_m\}$, and affine calibration terms $\{\gamma_m, \beta_m\}$. The primary objective is a soft-margin loss for binary evacuation prediction:
\begin{equation}
\mathcal{L}_{\text{task}} = \frac{1}{N}\sum_{i=1}^{N} \log\left(1 + \exp\left(-\tilde{y}_i \cdot \hat{z}(\mathbf{x}_i)\right)\right),
\end{equation}
where $\tilde{y}_i \in \{-1, +1\}$ is the label encoding. 

To prevent router collapse and encourage balanced expert utilization, we introduce auxiliary regularizers. A KL balance loss aligns the batch-averaged routing distribution $\bar{\boldsymbol{\pi}} = \frac{1}{N}\sum_{i=1}^{N} \boldsymbol{\pi}(\mathbf{x}_i)$ with the uniform prior $\mathcal{U}_M$.
\begin{equation}
\mathcal{L}_{\text{KL}} = \lambda_{\text{KL}} \cdot D_{\text{KL}}\left(\bar{\boldsymbol{\pi}} \,\|\, \mathcal{U}_M\right),
\end{equation}

An entropy regularizer maintains per-sample routing uncertainty,
\begin{equation}
\mathcal{L}_{\text{ent}} = -\lambda_{\text{ent}} \cdot \frac{1}{N}\sum_{i=1}^{N} H\left(\boldsymbol{\pi}(\mathbf{x}_i)\right).
\end{equation}

while a router z-loss stabilizes training by penalizing large router logits \citep{fedus2022switch}:
\begin{equation}
\mathcal{L}_{\text{z}} = \lambda_{\text{z}} \cdot \frac{1}{N}\sum_{i=1}^{N} \left(\log \sum_{m=1}^{M} \exp(a_m(\mathbf{x}_i))\right)^2,
\end{equation}
where $a_m(\mathbf{x})$ is the raw router logit for expert $m$.

Finally, we encourage expert diversity by penalizing squared cosine similarity between calibrated expert outputs \citep{shazeer2017outrageously}:
\begin{equation}
\mathcal{L}_{\text{div}} = \lambda_{\text{div}} \cdot \frac{1}{M^2}\sum_{m,m'} \left(\frac{\mathbf{g}_m^\top \mathbf{g}_{m'}}{\|\mathbf{g}_m\|\|\mathbf{g}_{m'}\|}\right)^2.
\end{equation}

The overall objective is
\begin{equation}
\mathcal{L} = \mathcal{L}_{\text{task}} + \mathcal{L}_{\text{KL}} + \mathcal{L}_{\text{ent}} + \mathcal{L}_{\text{z}} + \mathcal{L}_{\text{div}}.
\end{equation}

\paragraph{Training Stability.}

Stable training of symbolic MoE requires several practical techniques. We first use a router warm-up: during early epochs, routing weights are fixed to be uniform (or the router is frozen), preventing premature expert collapse before symbolic coefficients stabilize. KL balance regularization is activated only after this warm-up. To ensure numerical stability, we implement safe symbolic evaluation. Logarithm and square-root inputs are $\epsilon$-shifted; exponentials and power operations are clamped to bounded ranges; divisions are guarded against zero denominators; and all intermediate values are clipped to finite intervals, with NaN/Inf outputs replaced by zeros. Router logits are similarly clipped, and softmax stabilization is applied. We further employ gradient clipping, separate learning rates and weight decay for router and experts, optional router noise and expert dropout, and layer normalization in both router and coefficient networks. Training uses Adam with early stopping, best-checkpoint restoration, and inverse-frequency weighting to address class imbalance

\subsection{Baselines and Metrics}

We benchmark the proposed PASM framework against a diverse set of state-of-the-art models:

\noindent \textbf{XGBoost.} Gradient-boosted decision trees remain the dominant paradigm for tabular prediction tasks, consistently matching or outperforming neural models on medium-sized datasets \cite{chen2016xgboost, grinsztajn2022tree, rabbani2024attention}. We use XGBoost as the primary traditional ML benchmark.

\noindent \textbf{Large Language Models (GPT-5-mini medium reasoning efforts).} Recent studies apply LLMs to tabular prediction by prompting serialized features \cite{hegselmann2023tabllm, dinh2022lift}. While LLMs show promise in modeling human decision-making, they exhibit biases and limited robustness under distribution shift \cite{santurkar2023whose, vanRooij2024reclaiming}. We include GPT-5-mini with medium reasoning efforts to assess whether pretrained world knowledge improves cross-location evacuation prediction.

\noindent \textbf{TabPFN.} TabPFN \cite{grinsztajn2025tabpfn, hollmann2025tabpfn, hollmann2023tabpfn} is a transformer pretrained on synthetic tabular data to perform approximate Bayesian inference. Designed specifically for tabular classification, it provides a strong, low-tuning baseline and tests whether tabular-specific pretraining outperforms general-purpose models.

\paragraph{Evaluation Metrics.}
Evacuation data are highly imbalanced, especially in inland regions like Georgia, where “stay” decisions dominate, making accuracy alone misleading. A trivial always-stay classifier can achieve high accuracy while failing to identify evacuees.

We therefore use \textbf{Matthews Correlation Coefficient (MCC)} as the primary metric \cite{matthews1975comparison, boughorbel2017optimal}. MCC is equivalent to the Pearson correlation between predicted and true labels and is widely regarded as the most robust single metric for imbalanced binary classification. It rewards balanced performance across all confusion-matrix entries and ranges from $-1$ (perfect disagreement) to $+1$ (perfect agreement).
\begin{multline}
\text{MCC} = \\
\frac{\text{TP} \times \text{TN} - \text{FP} \times \text{FN}}{\sqrt{(\text{TP}+\text{FP})(\text{TP}+\text{FN})(\text{TN}+\text{FP})(\text{TN}+\text{FN})}}
\end{multline}
where TP, TN, FP, and FN denote true positives, true negatives, false positives, and false negatives, respectively. 

We additionally report \textbf{ROC-AUC} to assess threshold-independent ranking quality and \textbf{Accuracy} for completeness. All metrics are computed using standard scikit-learn implementations.
\begin{gather}
\text{ROC-AUC} = \int_0^1 \text{TPR}(t) \, d\text{FPR}(t) \\
\text{Accuracy} = \frac{\text{TP} + \text{TN}}{\text{TP} + \text{TN} + \text{FP} + \text{FN}}.
\end{gather}

\subsection{Implementation Details}

\paragraph{Symbolic Regression.}
All experiments were conducted on a workstation with an Intel i7-13700KF CPU and an NVIDIA RTX 4070 Ti GPU. The full pipeline was implemented in Python 3.13 using PyTorch 2.9.0. Symbolic regression was performed with the LaSR library \cite{grayeli2024lasr} (Julia 1.12.1), with LLM guidance provided by a locally deployed Qwen3-8B model via Ollama on an A6000 Ada server. Each run used 40 evolutionary generations with 30 parallel islands. The operator set included ${+, -, \times, /, \wedge}$ and ${\exp, \log, \sin, \cos, \sqrt{\cdot}}$, and expression trees were capped at 40 nodes. LLM-guided genetic operations were each assigned a trigger probability of 0.001. From the Pareto frontier balancing accuracy and complexity, we retained the top-5 expressions for the global expert and each cluster-specific expert.

\paragraph{MoE Architecture.}
The router and coefficient networks are MLPs with hidden dimension 128 and dropout 0.1, following a LayerNorm--Linear--ReLU--Dropout--Linear architecture. The coefficient network shares a backbone across experts with separate output heads. All parameters are trained jointly using Adam (learning rate $10^{-3}$) for up to 100 epochs, with early stopping (patience = 10) to prevent overfitting on the limited target-domain calibration data.

\paragraph{Computational Resources.}
The full symbolic regression pipeline required 2,068 LLM calls (4.5M tokens total) at the configured mutation probability of 0.001, completing in approximately 10 hours on the A6000 Ada GPU (131 tokens/s). Router and coefficient network training required only 45 seconds. Peak GPU memory was 5.7 GiB for the Ollama-hosted LLM and 1.6 GiB for router training.

\subsection{Data}\label{section:exp-data}

We evaluate PASM using the anonymized household survey dataset collected after Hurricanes Harvey (Texas) and Irma (Florida and Georgia) \cite{goodie2019experience, goodie2019data}. The dataset contains data from three feature domains: (1) demographic and socioeconomic attributes (age, sex, race, education, home ownership, household composition); (2) cognitive and experiential factors (risk perception, prior hurricane experience, past trauma, preparedness); and (3) social and environmental cues (the share of neighbors evacuated in the respondent’s ZIP code).

Our preliminary analysis reveals substantial population heterogeneity that limits cross-location generalization. To reflect realistic deployment, where models trained on past disasters inform future emergency management, we construct a source-target split. Florida and Texas form the source domain (Domain A), while Georgia serves as the target domain (Domain B). This split is motivated by an empirical test: models transfer reasonably well between Florida and Texas but degrade sharply in Georgia, which has a distinct demographic profile and a much higher evacuation rate (70\% vs.\ 33\%). Within Domain A, 85\% of samples are used to train symbolic experts, and 15\% are held out for validation. For MoE calibration, we augment the source validation set with a 100-shot random sample from Domain B. All remaining Georgia samples are reserved for final evaluation, with no hyperparameter tuning performed on the target test set. The dataset is publicly available on Mendeley Data \cite{goodie2019data}.

\section{Results}

\subsection{Symbolic MoE for Cross-location Adaptation}

Table~\ref{tab:main_results} summarizes performance on the Georgia test set. Higher values correspond to better performance. PASM achieves an MCC of 0.607, outperforming XGBoost (0.404), GPT-5-mini-medium (0.434), and TabPFN (0.333). This corresponds to relative improvements of 50\%, 40\%, and 82\%, respectively. Similar trends are observed on ROC-AUC: PASM achieves 0.840, compared to 0.680 for XGBoost, 0.692 for GPT-5-mini, and 0.751 for TabPFN. Overall, these results show that our proposed PASM framework transfers more reliably across states than existing tabular prediction approaches.

\begin{table}[htbp]
\centering
\small
\setlength{\tabcolsep}{4pt}
\begin{tabular}{l ccc}
\toprule
Model               & MCC            & ROC-AUC       & Accuracy\\
\midrule
XGBoost             & 0.404          & 0.680         & 0.692 \\
TabPFN              & 0.333          & 0.751         & 0.654 \\
GPT-5-mini(medium)  & 0.434          & 0.692         & 0.692 \\
\method             & \textbf{0.607} & \textbf{0.840} & \textbf{0.769} \\
\bottomrule
\end{tabular}
\caption{Cross-location adaptation performance in Georgia.}
\label{tab:main_results}
\end{table}

PASM also exhibits a smaller drop in performance when moving from calibration to held-out test data. On the calibration set (source validation plus 100 Georgia samples), it reaches MCC = 0.669, ROC-AUC = 0.921, and Accuracy = 0.833. When evaluated on the Georgia test set, MCC decreases to 0.607, a relative reduction of 9.3\%. In contrast, both XGBoost and TabPFN experience much larger performance losses when transferred across states, with MCC drops exceeding 55\% (Table~\ref{tab:cross_state_mcc}). This smaller gap indicates that the symbolic experts and the routing mechanism learn decision patterns that remain consistent across populations in different states. In practical terms, the model focuses on shared behavioral tendencies rather than state-specific quirks, making it better suited for real emergency planning, where insights from past disasters are expected to be applied to new locations.

\subsection{Comparison with Heterogeneity-Aware Methods}

To assess whether PASM's gains stem from its symbolic structure or simply from modeling heterogeneity, we compare against meta-learning methods designed for few-shot adaptation and heterogeneous data. Table~\ref{tab:meta_learning_baselines} reports results for MAML \cite{finn2017model}, Prototypical Networks \cite{snell2017prototypical}, Matching Networks \cite{vinyals2016matching}, and a hierarchical clustering baseline (HierClust+LR) that applies per-cluster logistic regression.

\begin{table}[htbp]
\centering
\small
\resizebox{\columnwidth}{!}{%
\setlength{\tabcolsep}{4pt}
\begin{tabular}{l ccc}
\toprule
Method & MCC & ROC-AUC & Accuracy \\
\midrule
Matching Networks & 0.280{\scriptsize$\pm$0.117} & 0.732{\scriptsize$\pm$0.096} & 0.589{\scriptsize$\pm$0.045} \\
MAML & 0.314{\scriptsize$\pm$0.158} & 0.753{\scriptsize$\pm$0.061} & 0.631{\scriptsize$\pm$0.071} \\
Prototypical Networks & 0.313{\scriptsize$\pm$0.162} & 0.763{\scriptsize$\pm$0.092} & 0.650{\scriptsize$\pm$0.078} \\
HierClust+LR & 0.346{\scriptsize$\pm$0.164} & 0.746{\scriptsize$\pm$0.100} & 0.635{\scriptsize$\pm$0.071} \\
\midrule
\method{} (Ours) & \textbf{0.607} & \textbf{0.840} & \textbf{0.769} \\
\bottomrule
\end{tabular}}
\caption{Comparison with meta-learning and heterogeneity-aware baselines.}
\label{tab:meta_learning_baselines}
\end{table}

PASM outperforms all meta-learning methods by substantial margins, with MCC improvements of +0.26 to +0.33. MAML learns shared initializations that adapt quickly to new tasks but does not produce interpretable subgroup-specific rules. Prototypical and Matching Networks operate in metric space without decomposing heterogeneity into distinct behavioral regimes. HierClust+LR is the closest structural analog, applying per-cluster models, but lacks the capacity of symbolic regression for nonlinear feature interactions. These results indicate that combining symbolic rule discovery with learned routing captures behavioral heterogeneity more effectively than adaptation-based approaches alone.

\paragraph{Discovered Symbolic Formulas.}
The routing mechanism assigns clusters to three distinct formula archetypes. The simplest archetype is a linear additive model, $\text{TimesAsked} + \text{EvacPctZip}$, routed to the youngest cohort (cluster 0, mean age 33.4). This formula encodes direct response to institutional and social pressure without nonlinear transformations. The most widely routed archetype balances geographic evacuation rate against logarithmic age resistance, $\text{EvacPctZip}/c_0 - \log(\text{Age})$, where the amplification factor $c_0 \approx 0.068$ magnifies community behavior approximately 14.7-fold. This formula serves five clusters (3, 4, 5, 6, 9) spanning diverse demographics. The most complex archetype integrates eight features with cosine and fourth-root transforms, modeling interactions between geographic signals, social isolation indicators (marital status), and demographic penalties (age, education). This multi-factor formula handles behaviorally complex groups, including those with non-standard marital status or extreme social isolation. These archetypes correspond to distinct decision mechanisms: direct social compliance, geographic pressure weighted against age-related resistance, and multi-factor risk integration. Full cluster profiling is provided in Appendix~\ref{sec:expert_profiles}.

\subsection{Ablation Studies}
We conduct two ablation studies to isolate the contributions of the key architectural components.

\paragraph{Effect of Learned Routing vs.\ Naive Aggregation.}
We first test whether the gains come from the symbolic experts themselves or from the learned routing. To disentangle these effects, we compare PASM to two simple aggregation baselines that do not use a router: (1) Top-1 Average, which applies the single best symbolic formula to all samples; and (2) Top-5 Average, which predicts using the average of the top-5 symbolic experts.

\begin{table}[htbp]
\centering
\small
\setlength{\tabcolsep}{4pt}
\begin{tabular}{l ccc}
\toprule
Method & MCC & ROC-AUC & Accuracy \\
\midrule
Top-1 Average & 0.234 & 0.766 & 0.615 \\
Top-5 Average & 0.548 & 0.769 & 0.731 \\
\midrule
No SR (LR + Router) & 0.434 & 0.752 & 0.731 \\
\midrule
\method & \textbf{0.607} & \textbf{0.840} & \textbf{0.769} \\
\bottomrule
\end{tabular}
\caption{Ablation on aggregation strategy.}
\label{tab:ablation_aggregation}
\end{table}

As shown in Table~\ref{tab:ablation_aggregation}, Top-1 Average performs poorly (MCC = 0.234), confirming that no single symbolic model generalizes across all target-domain subpopulations. Top-5 Average improves to MCC = 0.548 through ensembling, but PASM achieves a further 10.8\% relative gain (0.548 $\rightarrow$ 0.607) with a large improvement in ROC-AUC (0.769 $\rightarrow$ 0.840). Input-dependent routing, not static averaging, is the key to handling behavioral heterogeneity.

To test whether symbolic regression itself is necessary, we replace all symbolic experts with logistic regression models while keeping the router and coefficient adaptation intact. This variant achieves MCC = 0.434, a gain of 0.20 over Top-1 Average but 0.17 below full PASM. The gap shows that nonlinear feature interactions captured by symbolic formulas (cosine, log, and square-root transforms) encode decision mechanisms that linear models cannot represent.

\paragraph{Effect of Joint Coefficient Adaptation.}
We next test whether it is necessary to adapt expert coefficients jointly with the router, or whether simpler coefficient schemes are sufficient. We consider three variants: (1) Fixed Coefficients, where symbolic coefficients are frozen after symbolic regression, and only the router is trained; (2) Learnable (Static) Coefficients, where coefficients are trainable but shared across all inputs; and (3) PASM (Full), where both routing weights and coefficients are input-dependent and optimized jointly.

Table~\ref{tab:ablation_coefficients} reveals a counterintuitive result. Learnable (Static) coefficients perform worse than Fixed Coefficients in terms of MCC (0.365 vs. 0.389), despite introducing additional flexibility. This is plausible due to globally learned coefficients amplifying error propagation: a single set of coefficients must reconcile conflicting behavioral patterns across subpopulations, so updates that improve performance for one group can degrade it for another. These conflicts then compound through the routing mechanism.

\begin{table}[htbp]
\centering
\small
\setlength{\tabcolsep}{4pt}
\begin{tabular}{l ccc}
\toprule
Coefficient Strategy & MCC  & ROC-AUC & Accuracy \\
\midrule
Fixed Coefficients & 0.389 & 0.763 & 0.692 \\
Learnable (Static) & 0.365 & 0.828 & 0.654 \\
\method\ (Full) & \textbf{0.607} & \textbf{0.840} & \textbf{0.769} \\
\bottomrule
\end{tabular}
\caption{Ablation on coefficient adaptation strategy.}
\label{tab:ablation_coefficients}
\end{table}

In contrast, Fixed Coefficients keep expert behavior stable, allowing the router to focus on selecting the appropriate expert without interference from shifting coefficients. PASM resolves this trade-off by making the coefficients input dependent. Conditioning coefficients on household features allows each symbolic formula to adjust its response to different contexts. For example, applying different risk thresholds for residents in reinforced structures versus mobile homes. This joint, input-aware adaptation avoids cascading errors while enabling fine-grained personalization, leading to the strongest overall performance.

\section{Conclusion}
This paper introduced PASM, a population-adaptive symbolic mixture-of-experts framework for predicting evacuation decisions across states. Our analysis shows that cross-location generalization failures are not solely due to feature distribution shifts: even households with similar observable characteristics follow different decision patterns across three states. This behavioral heterogeneity makes single global models unreliable when deployed beyond their training region.

PASM addresses this by combining LLM-guided symbolic regression with a mixture-of-experts architecture. Symbolic models provide interpretable decision rules, while a learned router and input-dependent coefficient adaptation select and calibrate experts for different subpopulations. On the Georgia test set, using only 100 calibration samples, PASM achieves an MCC of 0.607, outperforming XGBoost (0.404), TabPFN (0.333), and GPT-5-mini (0.434) by 40--82\%. It also surpasses meta-learning baselines, including MAML (0.314), Prototypical Networks (0.313), and HierClust+LR (0.346), by MCC margins of +0.26 to +0.33, indicating that learned routing over symbolic experts captures behavioral heterogeneity more effectively than gradient-based or metric-space adaptation.

The routing mechanism discovers three formula archetypes: a two-variable linear model for the youngest cohort, a geographic-pressure formula weighted against age for middle-demographic clusters, and an eight-feature nonlinear formula for socially isolated groups. These archetypes produce interpretable behavioral profiles consistent with established evacuation sociology. A fairness audit across four demographic axes (race, sex, education, age) detects no statistically significant disparities after Bonferroni correction. Calibration experiments further show that 50 target-domain samples already recover over 90\% of the full-data MCC, suggesting that PASM can be deployed with minimal local data collection. Together, these properties make PASM a practical tool for emergency planning where interpretability and cross-region robustness are required.

\section*{Limitations}

PASM shows strong potential; however, several limitations remain. First, it is a “gray-box” model: although symbolic regression forms its core, the use of unsupervised subpopulation discovery and mixture-of-experts routing introduces elements that may reduce interpretability. Second, while inference is efficient, training is computationally intensive due to repeated LLM queries when fitting symbolic experts, making it more costly than standard tabular models. Third, the current UMAP + HDBSCAN clustering assumes discrete subpopulations, whereas real-world human heterogeneity is often continuous or hierarchical. Hard clustering may oversimplify fuzzy boundaries and overlapping memberships, while mathematically optimal partitions may not align with intuitive sociological categories, potentially reducing interpretability. A fairness audit across four demographic axes (race, sex, education, age) found no statistically significant disparities after Bonferroni correction, though a 4.7 percentage-point accuracy gap between male and female subgroups warrants continued monitoring (Appendix~\ref{sec:fairness}). The current evaluation is limited to US hurricane contexts in three states; adaptation to other cultural or geographic settings would require retraining the router on a local calibration sample and updating the symbolic regression concept library with region-specific domain knowledge. Future work will explore soft or probabilistic clustering and task-oriented clustering approaches that balance predictive performance with semantic interpretability.

\bibliography{custom}

@article{boughorbel2017optimal,
  author    = {Boughorbel, Sabri and Jarray, Fethi and El-Anbari, Mohammed},
  journal   = {PloS one},
  number    = {6},
  pages     = {e0177678},
  publisher = {Public Library of Science San Francisco, CA USA},
  title     = {Optimal classifier for imbalanced data using Matthews Correlation Coefficient metric},
  volume    = {12},
  year      = {2017}
}

@inproceedings{chen2016xgboost,
  author    = {Chen, Tianqi and Guestrin, Carlos},
  booktitle = {Proceedings of the 22nd ACM SIGKDD International Conference on Knowledge Discovery and Data Mining},
  doi       = {10.1145/2939672.2939785},
  pages     = {785--794},
  title     = {XGBoost: A Scalable Tree Boosting System},
  year      = {2016}
}

@article{chen2025symbolic,
  author  = {Chen, Justin Chih-Yao and Yun, Sukwon and Stengel-Eskin, Elias and Chen, Tianlong and Bansal, Mohit},
  journal = {arXiv preprint arXiv:2503.05641},
  title   = {Symbolic Mixture-of-Experts: Adaptive Skill-based Routing for Heterogeneous Reasoning},
  year    = {2025}
}

@article{Cranmer2020,
  author  = {Cranmer, Miles and Sanchez-Gonzalez, Alvaro and Battaglia, Peter and Xu, Rui and Cranmer, Kyle and Spergel, David and Ho, Shirley},
  journal = {Advances in Neural Information Processing Systems},
  title   = {Discovering Symbolic Models from Deep Learning with Inductive Biases},
  volume  = {33},
  year    = {2020}
}

@article{cutter2003social,
  author    = {Cutter, Susan L and Boruff, Bryan J and Shirley, W Lynn},
  journal   = {Social Science Quarterly},
  number    = {2},
  pages     = {242--261},
  publisher = {Wiley Online Library},
  title     = {Social vulnerability to environmental hazards},
  volume    = {84},
  year      = {2003}
}

@article{dai2024deepseekmoe,
  author  = {Dai, Damai and Deng, Chengqi and Zhao, Chenggang and Xu, RX and Gao, H and Chen, D and Li, J and Zeng, W and Yu, X and Wu, Y and others},
  journal = {arXiv preprint arXiv:2401.06066},
  title   = {Deepseekmoe: Towards ultimate expert specialization in mixture-of-experts language models},
  year    = {2024}
}

@article{dash2007evacuation,
  author    = {Dash, Nicole and Gladwin, Hugh},
  journal   = {Natural Hazards Review},
  number    = {3},
  pages     = {69--77},
  publisher = {American Society of Civil Engineers},
  title     = {Evacuation decision making and behavioral responses: Individual and household},
  volume    = {8},
  year      = {2007}
}

@article{dinh2022lift,
  author  = {Dinh, Tuan and Zeng, Yuchen and Zhang, Ruisu and Lin, Ziqian and Gira, Michael and Rajput, Shashank and Sohn, Jy-yong and Papailiopoulos, Dimitris and Lee, Kangwook},
  journal = {Advances in Neural Information Processing Systems},
  pages   = {11763--11784},
  title   = {Lift: Language-interfaced fine-tuning for non-language machine learning tasks},
  volume  = {35},
  year    = {2022}
}

@article{elder2007minorities,
  author    = {Elder, Keith and Xirasagar, Sudha and Miller, Nancy and Bowen, Shelly Ann and Glover, Saundra and Piper, Crystal},
  journal   = {American Journal of Public Health},
  number    = {S1},
  pages     = {S124--S129},
  publisher = {American Public Health Association},
  title     = {African {Americans}' decisions not to evacuate {New Orleans} before {Hurricane Katrina}: A qualitative study},
  volume    = {97},
  year      = {2007}
}

@article{fedus2022switch,
  author  = {Fedus, William and Zoph, Barret and Shazeer, Noam},
  journal = {Journal of Machine Learning Research},
  number  = {120},
  pages   = {1--39},
  title   = {Switch Transformers: Scaling to Trillion Parameter Models with Simple and Efficient Sparsity},
  volume  = {23},
  year    = {2022}
}

@article{fothergill2004poverty,
  author    = {Fothergill, Alice and Peek, Lori A},
  journal   = {Natural Hazards},
  number    = {1},
  pages     = {89--110},
  publisher = {Springer},
  title     = {Poverty and disasters in the {United States}: A review of recent sociological findings},
  volume    = {32},
  year      = {2004}
}

@article{gevaert2021fairnessdrm,
  author    = {Gevaert, Caroline M and Carman, Mary and Rosman, Benjamin and Georgiadou, Yola and Soden, Robert},
  journal   = {Patterns},
  number    = {11},
  publisher = {Elsevier},
  title     = {Fairness and accountability of AI in disaster risk management: Opportunities and challenges},
  volume    = {2},
  year      = {2021}
}

@article{goodie2019experience,
  author    = {Goodie, Adam S and Sankar, Adithya Raam and Doshi, Prashant},
  journal   = {International journal of disaster risk reduction},
  pages     = {101320},
  publisher = {Elsevier},
  title     = {Experience, risk, warnings, and demographics: Predictors of evacuation decisions in Hurricanes Harvey and Irma},
  volume    = {41},
  year      = {2019}
}

@article{grayeli2024lasr,
  author  = {Grayeli, Arya and Sehgal, Atharva and Costilla Reyes, Omar and Cranmer, Miles and Chaudhuri, Swarat},
  journal = {Advances in Neural Information Processing Systems},
  pages   = {44678--44709},
  title   = {Symbolic regression with a learned concept library},
  volume  = {37},
  year    = {2024}
}

@article{grinsztajn2022tree,
  author  = {Grinsztajn, L{\'e}o and Oyallon, Edouard and Varoquaux, Ga{\"e}l},
  journal = {Advances in neural information processing systems},
  pages   = {507--520},
  title   = {Why do tree-based models still outperform deep learning on typical tabular data?},
  volume  = {35},
  year    = {2022}
}

@article{hasan2011behavioral,
  author    = {Hasan, Samiul and Ukkusuri, Satish and Gladwin, Hugh and Murray-Tuite, Pamela},
  journal   = {Journal of Transportation Engineering},
  number    = {5},
  pages     = {341--348},
  publisher = {American Society of Civil Engineers},
  title     = {Behavioral model to understand household-level hurricane evacuation decision making},
  volume    = {137},
  year      = {2011}
}

@inproceedings{hegselmann2023tabllm,
  author       = {Hegselmann, Stefan and Buendia, Alejandro and Lang, Hunter and Agrawal, Monica and Jiang, Xiaoyi and Sontag, David},
  booktitle    = {International conference on artificial intelligence and statistics},
  organization = {PMLR},
  pages        = {5549--5581},
  title        = {Tabllm: Few-shot classification of tabular data with large language models},
  year         = {2023}
}

@misc{grinsztajn2025tabpfn,
  title         = {TabPFN-2.5: Advancing the State of the Art in Tabular Foundation Models},
  author        = {Léo Grinsztajn and Klemens Flöge and Oscar Key and Felix Birkel and Philipp Jund and Brendan Roof and
                   Benjamin Jäger and Dominik Safaric and Simone Alessi and Adrian Hayler and Mihir Manium and Rosen Yu and
                   Felix Jablonski and Shi Bin Hoo and Anurag Garg and Jake Robertson and Magnus Bühler and Vladyslav Moroshan and
                   Lennart Purucker and Clara Cornu and Lilly Charlotte Wehrhahn and Alessandro Bonetto and
                   Bernhard Schölkopf and Sauraj Gambhir and Noah Hollmann and Frank Hutter},
  year          = {2025},
  eprint        = {2511.08667},
  archiveprefix = {arXiv},
  url           = {https://arxiv.org/abs/2511.08667}
}

@article{hollmann2025tabpfn,
  title     = {Accurate predictions on small data with a tabular foundation model},
  author    = {Hollmann, Noah and M{\"u}ller, Samuel and Purucker, Lennart and
               Krishnakumar, Arjun and K{\"o}rfer, Max and Hoo, Shi Bin and
               Schirrmeister, Robin Tibor and Hutter, Frank},
  journal   = {Nature},
  year      = {2025},
  month     = {01},
  day       = {09},
  doi       = {10.1038/s41586-024-08328-6},
  publisher = {Springer Nature},
  url       = {https://www.nature.com/articles/s41586-024-08328-6}
}

@inproceedings{hollmann2023tabpfn,
  title     = {TabPFN: A transformer that solves small tabular classification problems in a second},
  author    = {Hollmann, Noah and M{\"u}ller, Samuel and Eggensperger, Katharina and Hutter, Frank},
  booktitle = {International Conference on Learning Representations 2023},
  year      = {2023}
}

@article{huang2016who,
  author    = {Huang, Shih-Kai and Lindell, Michael K and Prater, Carla S},
  journal   = {Environment and behavior},
  number    = {8},
  pages     = {991--1029},
  publisher = {Sage Publications Sage CA: Los Angeles, CA},
  title     = {Who leaves and who stays? A review and statistical meta-analysis of hurricane evacuation studies},
  volume    = {48},
  year      = {2016}
}

@article{jacobs1991adaptive,
  author    = {Jacobs, Robert A and Jordan, Michael I and Nowlan, Steven J and Hinton, Geoffrey E},
  journal   = {Neural computation},
  number    = {1},
  pages     = {79--87},
  publisher = {MIT Press},
  title     = {Adaptive mixtures of local experts},
  volume    = {3},
  year      = {1991}
}

@article{jiang2024mixtral,
  author  = {Jiang, Albert Q and Sablayrolles, Alexandre and Roux, Antoine and Mensch, Arthur and Savary, Blanche and Bamford, Chris and Chaplot, Devendra Singh and de las Casas, Diego and Hanna, Emma Bou and Bressand, Florian and others},
  journal = {arXiv preprint arXiv:2401.04088},
  title   = {Mixtral of experts},
  year    = {2024}
}

@article{lindell2012protective,
  author    = {Lindell, Michael K and Perry, Ronald W},
  journal   = {Risk Analysis: An International Journal},
  number    = {4},
  pages     = {616--632},
  publisher = {Wiley Online Library},
  title     = {The protective action decision model: Theoretical modifications and additional evidence},
  volume    = {32},
  year      = {2012}
}

@article{litman2006lessons,
  author    = {Litman, Todd},
  journal   = {Journal of Transportation Engineering},
  number    = {1},
  pages     = {11--18},
  publisher = {American Society of Civil Engineers},
  title     = {Lessons from {Katrina} and {Rita}: What major disasters can teach transportation planners},
  volume    = {132},
  year      = {2006}
}

@inproceedings{ma2024modula,
  author    = {Ma, Yufei and Liang, Zihan and Dai, Huangyu and Chen, Ben and Gao, Dehong and Ran, Zhuoran and Zihan, Wang and Jin, Linbo and Jiang, Wen and Zhang, Guannan and Cai, Xiaoyan and Yang, Libin},
  booktitle = {Proceedings of the 2024 Conference on Empirical Methods in Natural Language Processing},
  title     = {MoDULA: Mixture of Domain-Specific and Universal LoRA for Multi-Task Learning},
  url       = {https://aclanthology.org/2024.emnlp-main.161/},
  year      = {2024}
}

@article{matthews1975comparison,
  author    = {Matthews, Brian W},
  journal   = {Biochimica et Biophysica Acta (BBA) - Protein Structure},
  number    = {2},
  pages     = {442--451},
  publisher = {Elsevier},
  title     = {Comparison of the predicted and observed secondary structure of {T4} phage lysozyme},
  volume    = {405},
  year      = {1975}
}

@article{mcinnes2017hdbscan,
  title   = {hdbscan: Hierarchical density based clustering.},
  author  = {McInnes, Leland and Healy, John and Astels, Steve and others},
  journal = {J. Open Source Softw.},
  volume  = {2},
  number  = {11},
  pages   = {205},
  year    = {2017}
}

@article{mcinnes2018umap,
  title   = {UMAP: Uniform Manifold Approximation and Projection for Dimension Reduction},
  author  = {McInnes, Leland and Healy, John and Melville, James},
  journal = {stat},
  volume  = {1050},
  pages   = {6},
  year    = {2018}
}

@article{morss2016effects,
  title   = {The effects of past hurricane experiences on evacuation intentions through risk perception and efficacy beliefs: A mediation analysis},
  author  = {Demuth, Julie L and Morss, Rebecca E and Lazo, Jeffrey K and Trumbo, Craig},
  journal = {Weather, Climate, and Society},
  volume  = {8},
  number  = {4},
  pages   = {327--344},
  year    = {2016}
}

@incollection{perry2007disaster,
  author    = {Perry, Ronald W},
  booktitle = {Handbook of Disaster Research},
  editor    = {Rodr{\'i}guez, Havid{\'a}n and Quarantelli, Enrico L and Dynes, Russell R},
  pages     = {1--15},
  publisher = {Springer},
  title     = {What is a disaster?},
  year      = {2007}
}

@inproceedings{petersen2019deep,
  title     = {Deep symbolic regression: Recovering mathematical expressions from data via risk-seeking policy gradients},
  author    = {Brenden K Petersen and Mikel Landajuela Larma and Terrell N. Mundhenk and Claudio Prata Santiago and Soo Kyung Kim and Joanne Taery Kim},
  booktitle = {International Conference on Learning Representations},
  year      = {2021},
  url       = {https://openreview.net/forum?id=m5Qsh0kBQG}
}

@inproceedings{santurkar2023whose,
  title        = {Whose opinions do language models reflect?},
  author       = {Santurkar, Shibani and Durmus, Esin and Ladhak, Faisal and Lee, Cinoo and Liang, Percy and Hashimoto, Tatsunori},
  booktitle    = {International Conference on Machine Learning},
  pages        = {29971--30004},
  year         = {2023},
  organization = {PMLR}
}

@article{rabbani2024attention,
  author    = {Rabbani, Shourav B and Medri, Ivan V and Samad, Manar D},
  journal   = {International Journal of Data Science and Analytics},
  pages     = {1--23},
  publisher = {Springer},
  title     = {Attention versus contrastive learning of tabular data: a data-centric benchmarking},
  year      = {2024}
}

@article{schmidt2009distilling,
  author    = {Schmidt, Michael and Lipson, Hod},
  journal   = {science},
  number    = {5923},
  pages     = {81--85},
  publisher = {American Association for the Advancement of Science},
  title     = {Distilling free-form natural laws from experimental data},
  volume    = {324},
  year      = {2009}
}

@inproceedings{shazeer2017outrageously,
  author    = {Shazeer, Noam and Mirhoseini, Azalia and Maziarz, Krzysztof and Davis, Andy and Le, Quoc and Hinton, Geoffrey and Dean, Jeff},
  booktitle = {International Conference on Learning Representations},
  title     = {Outrageously large neural networks: The sparsely-gated mixture-of-experts layer},
  year      = {2017}
}

@inproceedings{shojaee2025llmsr,
  title     = {{LLM}-{SR}: Scientific Equation Discovery via Programming with Large Language Models},
  author    = {Parshin Shojaee and Kazem Meidani and Shashank Gupta and Amir Barati Farimani and Chandan K. Reddy},
  booktitle = {The Thirteenth International Conference on Learning Representations},
  year      = {2025},
  url       = {https://openreview.net/forum?id=m2nmp8P5in}
}

@inproceedings{standley2020tasks,
  title        = {Which tasks should be learned together in multi-task learning?},
  author       = {Standley, Trevor and Zamir, Amir and Chen, Dawn and Guibas, Leonidas and Malik, Jitendra and Savarese, Silvio},
  booktitle    = {International conference on machine learning},
  pages        = {9120--9132},
  year         = {2020},
  organization = {PMLR}
}

@article{sun2024predicting,
  title={Predicting hurricane evacuation decisions with interpretable machine learning methods},
  author={Sun, Yuran and Huang, Shih-Kai and Zhao, Xilei},
  journal={International Journal of Disaster Risk Science},
  volume={15},
  number={1},
  pages={134--148},
  year={2024},
  publisher={Springer}
}

@inproceedings{tian2023decompose,
  author    = {Tian, Zikang and Chen, Ruizhi and Hu, Xing and Li, Ling and Zhang, Rui and Wu, Fan and Peng, Shaohui and Guo, Jiaming and Du, Zidong and Guo, Qi and others},
  booktitle = {Advances in Neural Information Processing Systems},
  pages     = {66835--66858},
  title     = {Decompose a task into generalizable subtasks in multi-agent reinforcement learning},
  volume    = {36},
  year      = {2023}
}

@article{vanRooij2024reclaiming,
  author  = {van Rooij, Iris and Guest, Olivia and Adolfi, Federico G. and de Haan, Ronald and Kolokolova, Antonina and Rich, Patricia},
  journal = {Computational Brain \& Behavior},
  title   = {Reclaiming {AI} as a Theoretical Tool for Cognitive Science},
  volume  = {7},
  number  = {3},
  pages   = {343--356},
  year    = {2024},
  doi     = {10.1007/s42113-024-00217-5}
}

@article{wang2025drsr,
  title   = {DrSR: LLM based Scientific Equation Discovery with Dual Reasoning from Data and Experience},
  author  = {Wang, Runxiang and Wang, Boxiao and Li, Kai and Zhang, Yifan and Cheng, Jian},
  journal = {arXiv preprint arXiv:2506.04282},
  year    = {2025}
}

@article{Wong2023,
  author  = {Wong, Stephen D. and Broader, Jacquelyn C. and Walker, Joan L. and Shaheen, Susan A.},
  doi     = {10.1007/s11116-022-10275-y},
  journal = {Transportation},
  number  = {4},
  pages   = {1435--1473},
  title   = {Understanding California wildfire evacuee behavior and joint choice-making},
  volume  = {50},
  year    = {2023}
}

@inproceedings{yu2020gradient,
  author    = {Yu, Tianhe and Kumar, Saurabh and Gupta, Abhishek and Levine, Sergey and Hausman, Karol and Finn, Chelsea},
  booktitle = {Advances in Neural Information Processing Systems},
  pages     = {5824--5836},
  title     = {Gradient surgery for multi-task learning},
  volume    = {33},
  year      = {2020}
}

@inproceedings{zhao2025learning,
  author    = {Zhao, Zijie and Zhao, Zhongyue and Xu, Kaixuan and Fu, Yuqian and Chai, Jiajun and Zhu, Yuanheng and Zhao, Dongbin},
  booktitle = {Advances in Neural Information Processing Systems},
  title     = {Learning and Planning Multi-Agent Tasks via an MoE-based World Model},
  year      = {2025}
}

@article{zhao2025personalized,
  author  = {Zhao, Yibo and Zhao, Yang and Du, Hongru and Yang, Hao Frank},
  journal = {arXiv preprint arXiv:2511.02194},
  title   = {Personalized Decision Modeling: Utility Optimization or Textualized-Symbolic Reasoning},
  year    = {2025}
}

@misc{goodie2019data,
  title     = {Data for: Experience-Based and Demographic Predictors of Evacuation Decisions in Hurricanes Harvey and Irma},
  author    = {Goodie, Adam and Doshi, Prashant and Sankar, Adithya Raam},
  year      = {2019},
  publisher = {Mendeley Data},
  doi       = {10.17632/2cxnyxc8f8.1},
  url       = {https://data.mendeley.com/datasets/2cxnyxc8f8/1},
  version   = {1},
  note      = {Published: 2019-10-24}
}

@inproceedings{finn2017model,
  title={Model-agnostic meta-learning for fast adaptation of deep networks},
  author={Finn, Chelsea and Abbeel, Pieter and Levine, Sergey},
  booktitle={International Conference on Machine Learning},
  pages={1126--1135},
  year={2017},
  organization={PMLR}
}

@inproceedings{snell2017prototypical,
  title={Prototypical networks for few-shot learning},
  author={Snell, Jake and Swersky, Kevin and Zemel, Richard},
  booktitle={Advances in Neural Information Processing Systems},
  volume={30},
  year={2017}
}

@inproceedings{vinyals2016matching,
  title={Matching networks for one shot learning},
  author={Vinyals, Oriol and Blundell, Charles and Lillicrap, Timothy and Kavukciglu, Koray and Wierstra, Daan},
  booktitle={Advances in Neural Information Processing Systems},
  volume={29},
  year={2016}
}


\section{Appendix}
\label{sec:appendix}

\begin{figure*}[t]
\centering
\includegraphics[width=\linewidth]{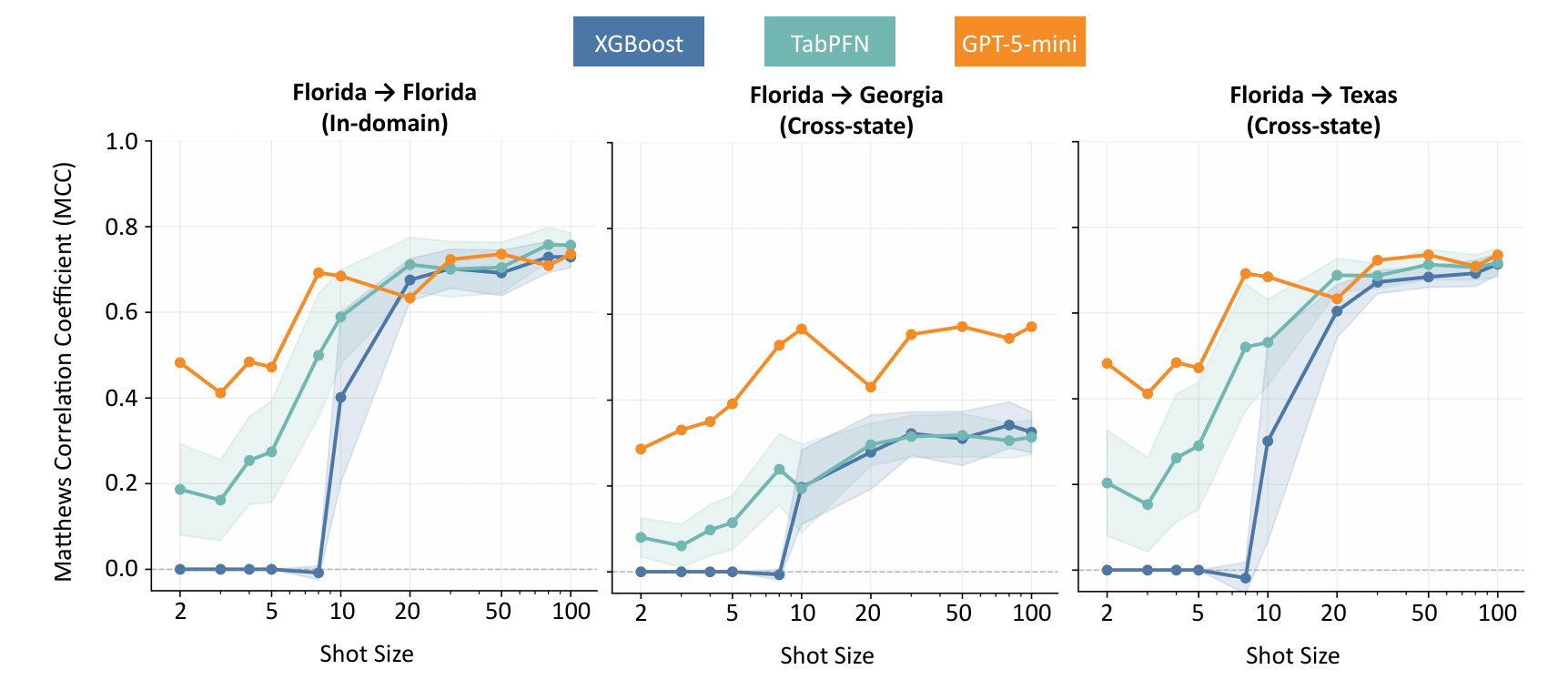}
\caption{Cross-state transferability: models trained on Florida and evaluated on Florida (intra-state) vs. Georgia/Texas (cross-state).}
\label{fig:stategap}
\end{figure*}

We conducted a preliminary analysis to reveal the existence and nature of the cross-state gap. These analyses provide the empirical justification for our architecture.

\subsection{Cross-State Transferability Gap}

We first examine cross-state performance degradation by training models on Florida data and evaluating them on other states under varying training sample sizes. As shown in Figure~\ref{fig:stategap}, all baseline models exhibit a substantial performance drop when transferred to other states, indicating a clear cross-state distribution shift. To quantify this cross-state transfer gap, Table~\ref{tab:cross_state_mcc} reports the Matthews correlation coefficient (MCC) for each model trained on 100 Florida samples and evaluated on test sets from all three states.

Several key patterns emerge. All models experience substantial performance degradation when evaluated on Georgia data: TabPFN declines from 0.757 to 0.313 ($-58.6\%$), XGBoost from 0.729 to 0.325 ($-55.4\%$), and GPT-5-mini from 0.736 to 0.571 ($-22.4\%$). In contrast, transferring models between Florida and Texas results in minimal performance loss. This asymmetric transfer behavior, where Florida and Texas generalize well to each other, but both perform poorly on Georgia, suggests the presence of latent population heterogeneity that is not captured by standard feature representations or conventional modeling approaches.

\begin{table*}[t]
\centering
\small
\setlength{\tabcolsep}{6pt}
\begin{tabular}{l ccc ccc ccc}
\toprule
& \multicolumn{3}{c}{Florida} & \multicolumn{3}{c}{Georgia} & \multicolumn{3}{c}{Texas} \\
\cmidrule(lr){2-4} \cmidrule(lr){5-7} \cmidrule(lr){8-10}
Model & MCC & AUC & Accuracy & MCC & AUC & Accuracy & MCC & AUC & Accuracy \\
\midrule
GPT-5-mini & .736 & .867 & .867 & .571 & .791 & .787 & .736 & .867 & .867 \\
TabPFN & .757{\scriptsize$\pm$.046} & .962{\scriptsize$\pm$.009} & .877{\scriptsize$\pm$.023} & .313{\scriptsize$\pm$.067} & .773{\scriptsize$\pm$.018} & .678{\scriptsize$\pm$.026} & .717{\scriptsize$\pm$.055} & .953{\scriptsize$\pm$.012} & .854{\scriptsize$\pm$.030} \\
XGBoost & .729{\scriptsize$\pm$.040} & .944{\scriptsize$\pm$.017} & .864{\scriptsize$\pm$.020} & .325{\scriptsize$\pm$.076} & .719{\scriptsize$\pm$.029} & .686{\scriptsize$\pm$.034} & .714{\scriptsize$\pm$.040} & .927{\scriptsize$\pm$.016} & .855{\scriptsize$\pm$.021} \\
\bottomrule
\end{tabular}
\caption{Cross-state transfer performance when training on Florida (100 shots) and testing on each state.}
\label{tab:cross_state_mcc}
\end{table*}

To illustrate the presence of state-specific effects and better understand their underlying mechanisms, we visualize evacuation probability differences using heatmaps projected onto a UMAP embedding space. Figure~\ref{fig:state_effect_heatmap} reveals a key insight: populations from different states exhibit distinct behavioral regimes. In the UMAP space, each point represents a fixed feature profile (e.g., identical age, income, and risk perception). If individuals across states followed the same decision function, the heatmap would be uniformly neutral, indicating no difference in predicted evacuation probabilities. Instead, the map displays pronounced spatial heterogeneity, with large red regions (probability differences of approximately +0.30 to +0.45) interspersed with localized blue patches.

This heterogeneous pattern has two important implications. First, the dominance of red regions indicates that, for most feature combinations, Georgia residents are more likely to evacuate than their counterparts in Florida or Texas. This points to a systematic “state effect”, potentially driven by differences in hurricane experience, state-level evacuation policies, media messaging, or community norms. Second, the presence of blue regions shows that for certain feature profiles, residents of Florida and Texas are more likely to evacuate, suggesting that the same decision factors (e.g., age or preparedness) can have opposite effects across states.

These findings have direct implications for model design. A single global model cannot simultaneously represent these conflicting behavioral patterns: learning one dominant relationship (e.g., higher preparedness increases evacuation likelihood) will inevitably misrepresent subpopulations in certain states. This observation motivates our Mixture-of-Experts approach, in which multiple experts capture distinct behavioral regimes and a learned routing mechanism dynamically combines them based on input features, enabling the model to adapt to cross-state behavioral heterogeneity.

\subsection{Heterogeneity Within a Target State}

Cross-state transfer is not the only challenge: heterogeneity among subpopulations within a single state can also limit the effectiveness of a unified global model. To examine this, we compare three training and testing configurations: (1) models trained on Florida data and evaluated on Florida (FL $\rightarrow$ FL), (2) models trained on Florida and evaluated on Georgia (FL $\rightarrow$ GA), and (3) models trained and evaluated on Georgia (GA $\rightarrow$ GA). If intra-state heterogeneity were minimal, we would expect configuration (3) to substantially outperform configuration (2), since both training and testing data would come from the same population.

Table~\ref{tab:intra_state_heterogeneity} shows a counterintuitive result: models trained on Georgia data (GA $\rightarrow$ GA) perform worse than Florida-trained models evaluated on Georgia (FL $\rightarrow$ GA). For TabPFN, the MCC decreases from 0.313 (FL $\rightarrow$ GA) to 0.274 (GA $\rightarrow$ GA), a relative decline of 12.5\%. XGBoost shows a more pronounced drop, from 0.325 to 0.224 ($-31.1\%$), while GPT-5-mini exhibits the largest degradation, with MCC falling from 0.571 to 0.295 ($-48.3\%$).

\begin{table}[htbp]
\centering
\small
\setlength{\tabcolsep}{3pt}
\begin{tabular}{l l ccc}
\toprule
Model & Train $\to$ Test & MCC & AUC & Accuracy \\
\midrule
\multirow{3}{*}{GPT-5-mini} 
  & FL $\to$ FL & .736 & .867 & .867 \\
  & FL $\to$ GA & .571 & .791 & .787 \\
  & GA $\to$ GA & .295 & .691 & .795 \\
\midrule
\multirow{3}{*}{TabPFN}
  & FL $\to$ FL & .757{\scriptsize$\pm$.046} & .962{\scriptsize$\pm$.009} & .877{\scriptsize$\pm$.023} \\
  & FL $\to$ GA & .313{\scriptsize$\pm$.067} & .773{\scriptsize$\pm$.018} & .678{\scriptsize$\pm$.026} \\
  & GA $\to$ GA & .274{\scriptsize$\pm$.101} & .763{\scriptsize$\pm$.066} & .729{\scriptsize$\pm$.037} \\
\midrule
\multirow{3}{*}{XGBoost}
  & FL $\to$ FL & .729{\scriptsize$\pm$.040} & .944{\scriptsize$\pm$.017} & .864{\scriptsize$\pm$.020} \\
  & FL $\to$ GA & .325{\scriptsize$\pm$.076} & .719{\scriptsize$\pm$.029} & .686{\scriptsize$\pm$.034} \\
  & GA $\to$ GA & .224{\scriptsize$\pm$.091} & .694{\scriptsize$\pm$.073} & .672{\scriptsize$\pm$.025} \\
\bottomrule
\end{tabular}
\caption{Within-state heterogeneity analysis (100 training shots).}
\label{tab:intra_state_heterogeneity}
\end{table}

This seemingly paradoxical outcome, where out-of-domain training data generalize better than in-domain data, can be explained by the interaction between intra-state population heterogeneity and limited training samples. Georgia’s population likely consists of multiple latent subgroups with distinct evacuation decision patterns. With a small training set (100 shots), models trained solely on Georgia data may overfit to the specific subgroups represented in the sample, failing to capture the broader behavioral diversity of the state. In contrast, Florida’s larger and more diverse training distribution may induce more robust decision boundaries that transfer better to Georgia’s heterogeneous population, despite originating from a different state.

These findings reinforce and extend the earlier cross-state transferability analysis (Figure~\ref{fig:stategap}). The observed transfer gap is not solely a cross-state phenomenon but reflects a deeper structural challenge: behavioral heterogeneity operates at multiple levels, both across states and within individual states. As a result, a single global model, whether trained on the source or target domain, tends to favor dominant behavioral patterns while underperforming for minority subgroups. This multi-level heterogeneity further motivates the proposed Mixture-of-Experts approach, which can identify and specialize in distinct behavioral regimes regardless of their geographic origin.

\subsection{Theory-Inspired Two-Stage Prediction Still Faces a Transfer Gap}

The Protective Action Decision Model (PADM) is a foundational theoretical framework for evacuation behavior analysis. It conceptualizes disaster decision-making as a sequence of interpretable cognitive stages, progressing from hazard cues to risk perception and from risk perception to protective action \cite{lindell2012protective, huang2016who}. This two-stage decomposition offers strong theoretical grounding and has informed a wide range of empirical studies \cite{morss2016effects, hasan2011behavioral}. Motivated by this structure, we examine whether a PADM-inspired architecture can improve cross-state generalization.

Specifically, we evaluate an oracle setting in which the second-stage evacuation decision model is provided with ground-truth risk perception values rather than predictions from the first stage, thereby eliminating error propagation from perception modeling. If PADM’s cognitive decomposition captures the fundamental and transferable structure of evacuation decision-making, this oracle configuration should substantially reduce cross-state performance degradation. 

\begin{table}[htbp]
\centering
\small
\setlength{\tabcolsep}{4pt}
\begin{tabular}{l l ccc}
\toprule
Model & Perception & MCC & AUC & Acc \\
\midrule
\multirow{2}{*}{GPT-5-mini}
  & Without & .571 & .791 & .787 \\
  & Oracle & .629 & .822 & .813 \\
\midrule
\multirow{2}{*}{TabPFN}
  & Without & .313{\scriptsize$\pm$.067} & .773{\scriptsize$\pm$.018} & .678{\scriptsize$\pm$.026} \\
  & Oracle & .335{\scriptsize$\pm$.055} & .797{\scriptsize$\pm$.021} & .694{\scriptsize$\pm$.021} \\
\midrule
\multirow{2}{*}{XGBoost}
  & Without & .325{\scriptsize$\pm$.076} & .719{\scriptsize$\pm$.029} & .686{\scriptsize$\pm$.034} \\
  & Oracle & .349{\scriptsize$\pm$.038} & .765{\scriptsize$\pm$.024} & .699{\scriptsize$\pm$.018} \\
\bottomrule
\end{tabular}

\caption{PADM-inspired two-stage prediction: Florida$\to$Georgia transfer (100 shots).}
\label{tab:padm_oracle}
\end{table}

However, Table~\ref{tab:padm_oracle} shows that cross-state transfer remains limited even under this oracle condition. For TabPFN, the MCC increases only marginally from 0.313 to 0.335 (+7.0\%), and for XGBoost from 0.325 to 0.349 (+7.4\%). GPT-5-mini exhibits a larger improvement, from 0.571 to 0.629 (+10.2\%), yet still falls well below within-state performance levels (Table~\ref{tab:cross_state_mcc}, where Florida MCC exceeds 0.73).

These results indicate that cross-state generalization challenges persist beyond errors in modeling risk perception. Differences across states likely arise not only in how residents form risk perceptions from identical hazard cues, but also in how similar perceptions are translated into evacuation decisions. Such variation may reflect state-specific factors, including prior hazard experience, cultural norms, institutional practices, or infrastructure constraints.

\subsection{Symbolic Expert Cluster Profiles}
\label{sec:expert_profiles}

The routing mechanism assigns each cluster to a symbolic expert based on learned activation weights. Table~\ref{tab:cluster_profiles} summarizes the 10 discovered clusters, showing the routed expert formula, key demographic features, and behavioral interpretation for each.

\begin{table*}[t]
\centering
\small
\setlength{\tabcolsep}{4pt}
\begin{tabular}{c l p{4.5cm} p{3.5cm}}
\toprule
Cluster & Expert Formula & Demographic Profile & Behavioral Pattern \\
\midrule
C0 & $\text{TimesAsked} + \text{EvacPctZip}$ & Youngest (age 33.4), unmarried, low education & Direct response to external pressure \\
C1 & $\frac{\text{EvacPctZip}}{c_0} + \ldots - \log(\text{Age})$ & High age (69.5), non-standard marital status & Geographic signal vs. age resistance \\
C2 & $\frac{\text{EvacPctZip}}{c_0} + \ldots - \log(\text{Age})$ & Near-mean demographics, married females & Calibration baseline group \\
C3 & $\frac{\text{EvacPctZip}}{c_0} - \log(\text{Age})$ & Older (59.0), widowed females & Community behavior vs. age \\
C4 & $\frac{\text{EvacPctZip}}{c_0} - \log(\text{Age})$ & Young (34.5), divorced, low education & Geographic rate vs. age \\
C5 & $\frac{\text{EvacPctZip}}{c_0} - \log(\text{Age})$ & Young (33.1), high education, males with children & Minimal age resistance, high compliance (96.9\%) \\
C6 & $\frac{\text{EvacPctZip}}{c_0} - \log(\text{Age})$ & Older (64.5), married males, high external pressure & Strong community signal (75.5\%) \\
C7 & $\frac{\text{EvacPctZip}}{c_0} + \ldots - \log(\text{Age})$ & Older (60.3), high education, socially isolated & Low external cues, education offset \\
C8 & $\frac{\text{EvacPctZip}}{c_0} + \ldots - \log(\text{Age})$ & Older (60.1), married males, extreme isolation & Zero evacuation (0.0\%) \\
C9 & $\frac{\text{EvacPctZip}}{c_0} - \log(\text{Age})$ & Older (63.1), high education, multi-story homes & Structural protection perception \\
\bottomrule
\end{tabular}
\caption{Cluster profiles showing routed expert formulas, demographic characteristics, and behavioral interpretations. Three formula archetypes emerge: (A) linear additive (C0 only), (B) geographic rate vs. log-age (C3,4,5,6,9), and (C) multi-factor with cosine and root transforms (C1,2,7,8).}
\label{tab:cluster_profiles}
\end{table*}

Three distinct formula archetypes emerge from the routing analysis. Formula A (Expert 25, Cluster 0 only) uses a simple linear sum of $\text{TimesAsked}$ and $\text{EvacPctZip}$, capturing direct response to external pressure without nonlinear transformations. Formula B (Expert 26, Clusters 3,4,5,6,9) implements a two-term tradeoff between amplified geographic evacuation rate ($\text{EvacPctZip}/c_0$, with amplification factor approximately 14.7) and logarithmic age resistance. Formula C (Expert 3, Clusters 1,2,7,8) integrates eight features through cosine and fourth-root transforms, modeling complex interactions between geographic signals, social isolation indicators, and demographic penalties.

The routing mechanism assigns semantically coherent clusters to formula archetypes. The simplest social-pressure formula serves the youngest cohort (C0, mean age 33.4). Age-resistance formulas serve mid-to-older populations across multiple clusters with varying demographic contexts. Multi-factor formulas handle behaviorally complex groups, including those with non-standard marital status (C1), near-mean demographics requiring fine-grained calibration (C2), or extreme social isolation (C7, C8).

\subsection{Clustering Stability Analysis}
\label{sec:clustering_stability}

To verify that the discovered subpopulations are not artifacts of random initialization, we evaluate clustering stability under two sources of randomness: UMAP embedding seed variation (Experiment A) and full pipeline randomness including data-split variation (Experiment B).

\begin{table}[htbp]
\centering
\small
\resizebox{\columnwidth}{!}{%
\setlength{\tabcolsep}{6pt}
\begin{tabular}{l c c}
\toprule
Metric & Value & Threshold \\
\midrule
Adjusted Rand Index (ARI) & 0.876 & $>$0.8 (strong) \\
Normalized Mutual Information (NMI) & 0.916 & $>$0.8 (high) \\
Co-clustering Jaccard & 0.822 & $>$0.75 (reliable) \\
Cluster count (mode) & 8 & range [6,11] \\
Noise fraction & 0.020 & --- \\
\bottomrule
\end{tabular}}
\caption{Clustering stability metrics for Experiment A (20 runs with varying UMAP random state, fixed data split).}
\label{tab:clustering_stability}
\end{table}

Table~\ref{tab:clustering_stability} reports permutation-invariant stability metrics for Experiment A, where UMAP random state varies across 20 runs while the data split remains fixed. The high ARI (0.876) and NMI (0.916) values confirm that cluster assignments remain consistent across random seeds, with over 82\% pairwise co-clustering agreement (Jaccard).

Experiment B tests a stronger perturbation: both the data-split seed and UMAP initialization vary across 10 independent runs, so the training sample composition changes in each run. Despite this additional source of variation, structural properties remain stable: the cluster count concentrates at mode 6 (range [5, 10]) with mean $6.6 \pm 1.5$, and the noise fraction averages $0.014 \pm 0.022$, confirming that the vast majority of samples are assigned to well-defined density regions regardless of the specific training fold. These results demonstrate that the discovered subpopulations reflect stable density structures in the representation space rather than initialization artifacts.

\subsection{Computational Cost and Ablation Details}
\label{sec:cost_breakdown}

Table~\ref{tab:cost_breakdown} reports computational costs for \method{} and two ablation baselines. The LLM is invoked only during symbolic regression search with probability $p=0.001$ per genetic operation, not during router training or inference. The full pipeline completes within a single workday on commodity GPU hardware. Once experts are discovered, deployment requires only the lightweight router (45 seconds to train, negligible inference time), making the one-time symbolic search cost acceptable for practical applications.

\begin{table}[htbp]
\centering
\small
\resizebox{\columnwidth}{!}{%
\setlength{\tabcolsep}{4pt}
\begin{tabular}{l c c c c}
\toprule
Method & MCC & LLM Calls & LLM Tokens & Wall Time (s) \\
\midrule
\method{} (Full) & 0.607 & 2,068 & 4,508,756 & 24,699 \\
No Routing (Top-1 SR) & 0.234 & 150 & 313,483 & 1,718 \\
No SR (LR + Router) & 0.434 & 0 & 0 & 45 \\
\bottomrule
\end{tabular}}
\caption{Computational cost breakdown. GPU memory: Ollama LLM server 5.720 GiB peak, router training 1,576 MiB. Throughput: 131 tok/s (A6000 Ada), 79 tok/s (RTX 4070 Ti).}
\label{tab:cost_breakdown}
\end{table}

\subsection{Calibration Sample-Size Sensitivity}
\label{sec:sample_sensitivity}

We ablate MoE router training on varying numbers of target-domain calibration samples (20, 30, 50, 80, 100), fixing the source-domain symbolic experts and the Georgia test set. Table~\ref{tab:sample_sensitivity} reports MCC for each sample size.

\begin{table}[htbp]
\centering
\small
\setlength{\tabcolsep}{5pt}
\begin{tabular}{c c}
\toprule
Calibration Shots & MCC \\
\midrule
20 & 0.426 \\
30 & 0.488 \\
50 & 0.548 \\
80 & 0.566 \\
100 & 0.607 \\
\bottomrule
\end{tabular}
\caption{MoE router calibration MCC across target-domain sample sizes. Source-domain experts and Georgia test set are held fixed.}
\label{tab:sample_sensitivity}
\end{table}

MCC rises monotonically from 0.426 (20 shots) to 0.548 (50 shots) and 0.607 (100 shots). The 20-to-50 gain (+0.12) roughly doubles the 50-to-100 gain (+0.06), indicating diminishing returns. Thus 50 calibration samples already yield strong router performance, and quantitative benefits plateau beyond 100.

\subsection{Policy-Relevant Demographic Grid Comparison}
\label{sec:demographic_grid}

To assess the tradeoff between interpretability and predictive performance, we compare the data-driven UMAP+HDBSCAN clustering (main pipeline) against a policy-relevant demographic grid defined by Age (young/middle/old) $\times$ Education (high/low), yielding 6 groups. The demographic grid achieves MCC = 0.457 $\pm$ 0.086, compared to \method{} (data-driven) MCC = 0.607, a gap of 0.15.

This gap reflects a fundamental tradeoff. Demographic categories align with policy-relevant groupings (e.g., "elderly with low education") but miss latent behavioral heterogeneity that crosses demographic boundaries. Data-driven clustering captures these cross-cutting patterns at the cost of less intuitive group labels. A hybrid approach, using demographic priors as initialization for representation learning, may combine the strengths of both strategies by preserving interpretability while adapting to behavioral structure.

\subsection{Demographic Fairness Analysis}\label{sec:fairness}

We evaluate prediction fairness across four demographic axes (race, sex, education, age) using Fisher exact tests with Bonferroni correction. Table~\ref{tab:fairness} reports accuracy for each group and statistical significance of differences. No axis shows a statistically significant accuracy disparity after correction (all corrected $p > 0.05$). The largest effect appears on the sex axis, where male respondents receive 4.7 percentage points higher accuracy than female respondents, but this difference does not reach significance (corrected $p = 0.259$). We flag this gap for continued monitoring in future deployments.

\begin{strip}
\captionsetup{hypcap=false}
\centering
\small
\setlength{\tabcolsep}{5pt}
\captionof{table}{Demographic fairness analysis (N=679, overall accuracy 88.5\%). No statistically significant disparities detected after Bonferroni correction.}
\label{tab:fairness}
\vspace{2pt}
\begin{tabular}{l l l c c c c}
\toprule
Axis & Group 0 & Group 1 & Acc(G$_0$) & Acc(G$_1$) & $\Delta$ Acc & Fisher $p$ (Bonf.) \\
\midrule
Race & Non-White & White & 0.897 & 0.880 & +0.016 & 1.000 \\
Sex & Female & Male & 0.867 & 0.914 & $-$0.047 & 0.259 \\
Education & High & Low & 0.870 & 0.907 & $-$0.037 & 0.578 \\
Age & Old & Young & 0.899 & 0.871 & +0.027 & 1.000 \\
\bottomrule
\end{tabular}
\end{strip}


\end{document}